\RequirePackage{lineno}
\documentclass[twocolumn,pre,floatfix,superscriptaddress,amsmath,amssymb,nobalancelastpage]{revtex4}

\usepackage{nameref}
\usepackage{hyperref}
\usepackage{graphicx}
\usepackage{pstricks}
\usepackage[T1]{fontenc}
\usepackage{float}
\usepackage{dsfont}
\usepackage{natbib}
\usepackage{appendix}
\usepackage{amsmath} 

\usepackage{blindtext}

\begin{document}
\title{
Unsupervised detection of semantic correlations in big data
}
\begin{abstract}
In real-world data, information is stored in extremely large feature vectors. These variables are typically correlated due to complex interactions involving many features simultaneously. Such correlations qualitatively correspond to semantic roles and are naturally recognized by both the human brain and artificial neural networks. This recognition enables, for instance, the prediction of missing parts of an image or text based on their context.
We present a method to detect these correlations in high-dimensional data represented as binary numbers. We  estimate the \textit{binary intrinsic dimension} of a dataset, which quantifies the minimum number of independent coordinates needed to describe the data, and is therefore a proxy of semantic complexity.
The proposed algorithm is largely insensitive to the so-called curse of dimensionality, and can therefore be used in big data analysis. 
We test this approach identifying phase transitions in model magnetic systems and we then apply it to the detection of semantic correlations of images and text inside deep neural networks.

\end{abstract}

\author{Santiago Acevedo}
\email{sacevedo@sissa.it}
\affiliation{International School for Advanced Studies (SISSA), Via Bonomea 265,  34136 Trieste, Italy}
\author{Alex Rodriguez}
\email{alejandro.rodriguezgarcia@units.it}
\affiliation{Dipartimento di Matematica e Geoscienze, Università degli Studi di Trieste, via Alfonso Valerio 12/1, 34127 Trieste, Italy}
\affiliation{The Abdus Salam International Centre for Theoretical Physics (ICTP), Strada Costiera 11, 34014 Trieste, Italy} 
\author{Alessandro Laio}
\email{laio@sissa.it}
\affiliation{International School for Advanced Studies (SISSA), Via Bonomea 265,  34136 Trieste, Italy}
\affiliation{The Abdus Salam International Centre for Theoretical Physics (ICTP), Strada Costiera 11, 34014 Trieste, Italy} 
\maketitle

\section*{Introduction}

Correlations among the features defining a data point force a data set to live in a manifold $\mathcal{M}$ whose geometry encodes the correlation structure of data. The most fundamental property of $\mathcal{M}$ is its intrinsic dimension\cite{Bennett_intrinsic} (ID) which is qualitatively defined by the number of independent variables needed to parameterize it without significant information loss. 
If one was able to estimate the intrinsic dimension in a generic big data scenario, one would be able to quantify the strength of correlations in real-world datasets, like a corpus of sentences or a large set of measurements from a camera. However, real-world data are represented in feature spaces that are outrageously large, and most of the available methods significantly struggle if the intrinsic dimension of the manifold is of the order of hundreds or more \cite{camastra2016Review}. 
Detecting and quantifying semantic correlations in generic real world data is a complex task since features live in high dimensional spaces and have interactions of arbitrary range and nature. For example, in an image, the meaning of a scene can emerge from the simultaneous presence of two objects that are physically far away. Moreover, semantic correlations are often associated with collective effects, such as the simultaneous presence of three or more specific objects. A prominent example where non-trivial correlations emerge is the feature space of deep neural networks. Those spaces typically lack a well-defined order, and neurons that are close in the architecture are not necessarily coding related information. However, the activations are known to be contained in a manifold of relatively low dimension\cite{goldt2020modeling}, and the existence of this manifold is believed to be an important ingredient in the learning process. 
The main idea of this work is to use the intrinsic dimension (ID) of a dataset to detect and quantify correlations. If, for example, in a long text correlation between the first and the second part of the text vanish, the ID of the whole text should be approximately equal to the sum of the ID of the two parts. More precisely, in data points including features describing independent information, the ID should scale linearly with the number of features N, namely $ID = \gamma N$ for some positive constant $\gamma$. If one were able to estimate the ID as a function of text length (or as a function of the patch dimension, in the case of images) one would be able to detect the range of the correlation. Moreover, the difference between the sum of the IDs of two representations and the ID of a dataset composed by the concatenation of these representations provides a quantitative measure of the strength of the correlations\cite{basile2024intrinsic}: for perfectly correlated representations the ID of the concatenation will be equal to the ID of the single representations. 
However, most of the available algorithms to perform ID estimation meet serious problems if the intrinsic dimension of the manifold is of the order of hundreds or more\cite{camastra2016Review}. In particular, some of them fail to provide an estimate of the ID which scales linearly with the number of features even when those are generated independently. Indeed, methods relying on the statistics of local neighborhoods systematically underestimate the ID because the number of samples required to sample a dataset grows exponentially in the ID\cite{ECKMANN1992185}. This problem is a manifestation of the so-called curse of dimensionality\cite{bishop2016pattern} and affects to different extent all the ID estimators we are aware of, even if recently some progress has been made\cite{erba2019intrinsic}.

Binary variables are the basis of all digital computation but are also of fundamental importance in both theoretical and applied science. The so-called spin systems, in which a spin is a binary degree of freedom, provide standard models to understand phases of matter, such as spin glasses\cite{spin_glasses_review_binder}, and spin liquids\cite{spin_liquids_review}, and the corresponding phase transitions, like spontaneous ordering\cite{kardar2007statistical}. Binary variables are also used to model social dynamics\cite{ising_social}, biological neurons\cite{hopfield1982neural} and learning processes\cite{engel2001statistical,barbier2025spins,Marino_2024}. In artificial neural networks, the activations are typically real numbers, but many efforts focus on finding more efficient representations: the so-called quantization methods\cite{hubara2018quantized,guo2018survey} aim to represent activations or weights using small integer numbers or binary variables, in order to drastically save memory, energy, and computing time, while preserving performance. Activations and weights can be transformed in binary variables by the sign function\cite{courbariaux2016binarized,wang2023bitnet}, which has been shown to preserve the scalar product between them\cite{anderson2017high}. 
Here, we introduce an intrinsic dimension estimator that generalizes the concept of the intrinsic dimension to the case of binary variables. We present an algorithm that can be used to estimate the binary intrinsic dimension (BID) of very large datasets of binary variables.  We first show that it provides estimates that scale approximately linearly with system size when the system can be split into subsystems that are statistically independent. Concretely, we test the algorithm up to one million spins, using only 2500 samples. Then we benchmark the algorithm on model systems in which the binary variables interact with an explicit energy function, which allows modulating their correlations. We show that the dependence of the BID on system size allows for characterizing the nature of correlations between the variables and spotting phase  transitions. In Supplementary Note 4 We show that binarization also approximately preserves the local neighbourhoods computed using the full-precision activations inside a Large Language Model (LLM). This suggests that binarizing in high dimensions is an operation that retains relevant information about the data\cite{inf-imb}. We finally apply this approach to the study of the internal representations of images and text in large neural networks. This result paves the way to a plethora of applications, allowing to address questions which have thus far been, to the best of our knowledge, intractable. Our approach is effective in a variety of fields and topics, from the study of phase transitions and of frustration-induced disorder in statistical physics, to image classification.

\section*{Results}
\label{sec:Results} 

\subsection*{Theoretical model}
\label{sec:binomial-estimator}

Our algorithm is based on a generalization of exact results for distance distributions between statistically independent binary variables.  
Let $\boldsymbol{\sigma} \in \mathcal{C} = \{ 0,1 \}^N$ be a string of $N$ independent bits, and let each bit be uniformly distributed in $\{0,1\}$. Then, the probability of observing a Hamming distance $r$ between two configurations $\boldsymbol{\sigma}$ and $\boldsymbol{\sigma}'$ is 

\begin{equation}
P_0(r) 
= \binom{N}{r} p^r ( 1 - p )^{N-r},
\label{eq:p-of-d-free}
\end{equation}
with $p=1/2$, which can take the compact form $P_0(r)=\frac{1}{2^N} \binom{N}{r}$.
Geometrically, the coordinates of uncorrelated random bits are points uniformly distributed in the $N$-dimensional configuration space $\mathcal{C}$, the corners of a high dimensional hypercube, and the intrinsic dimension $d$ coincides with the number of variables, $N$. 
In the case of correlated spins, we assume that the dimension is a smooth function of the distance $r$, and therefore make the following ansatz:
\begin{equation}
P(r)=C\frac{1}{2^{d(r)}}
\binom{d(r)}{r},
\label{eq:p-of-d}
\end{equation}
where $C$ is a normalization constant. Eq. \eqref{eq:p-of-d} reduces to \eqref{eq:p-of-d-free}
for $ d(r) = N$ and $C=1$. We empirically found that model \eqref{eq:p-of-d} fits accurately the observed distributions, at least locally, if one retains the first two terms of the Taylor expansion:
\begin{equation}
    d(r)=d_0+d_1 \, r
    \label{eq:d-of-r}
\end{equation}
where $d_0$ and $d_1$ are variational parameters.
In Methods, section `Maximum Likelihood Estimation of a scale-dependent BID' we show a Maximum Likelihood Estimation approach to infer a scale-dependent BID that supports our assumption of  linear dependence \eqref{eq:d-of-r}.

In order to infer $d_0$ and $d_1$
we minimize the Kullbac-Leibler divergence between the empirical probability of Hamming distances $P_{emp}(r)$ and the model $P(r)$ given by Eqs. \eqref{eq:p-of-d} and \eqref{eq:d-of-r}, 

\begin{equation}
    D_{KL}(P_{emp}||P) = \sum_{r \leq r^*} P_{emp}(r) \log{\frac{P_{emp}(r)}{P(r)}},
    \label{eq:KLdiv}
\end{equation}
where $r^*$ is a meta-parameter that allows constraining the fit to small distances (see Supplementary Note 1). 
As a representative measure of the BID of a system, we consider the limit for small distances of $d(r)$, which qualitatively corresponds to the local number of degrees of freedom. This definition is consistent with the idea that, if the data are defined by real variables, the ID should be equal to the dimension of the hyperplane that is tangent to the data manifold, which can be estimated considering only small distances. According to Eq. \eqref{eq:d-of-r}, we simply have $BID \equiv d_0$. 
The computational complexity of computing our observable $P_{emp}(r)$ scales linearly with the number of features and quadratically with the number of samples, since it involves computing all the  Hamming distances between each pair of points. \\
Eq. \eqref{eq:KLdiv} allows us to quantify rigorously how well our model approximates the observed data, $P_{emp}(r)$.
In order to assign an uncertainty to the value obtained by the optimization, one can perform Bayesian inference with a Markov Chain Monte Carlo (MCMC) on the parameter space of model \eqref{eq:p-of-d}, given the observations.
In Methods, section `Bayesian inference of the BID by Markov Chain Monte Carlo'  we show that both the minimization of Eq. \eqref{eq:KLdiv} and MCMC Bayesian inference with a uniform prior provide similar results. MCMC allows estimating the posterior distribution of two parameters of the model, and, as a consequence their mean and variances.



\subsection*{Numerical experiments}

\begin{figure*}[ht!]
\centering
\includegraphics[width=\linewidth]{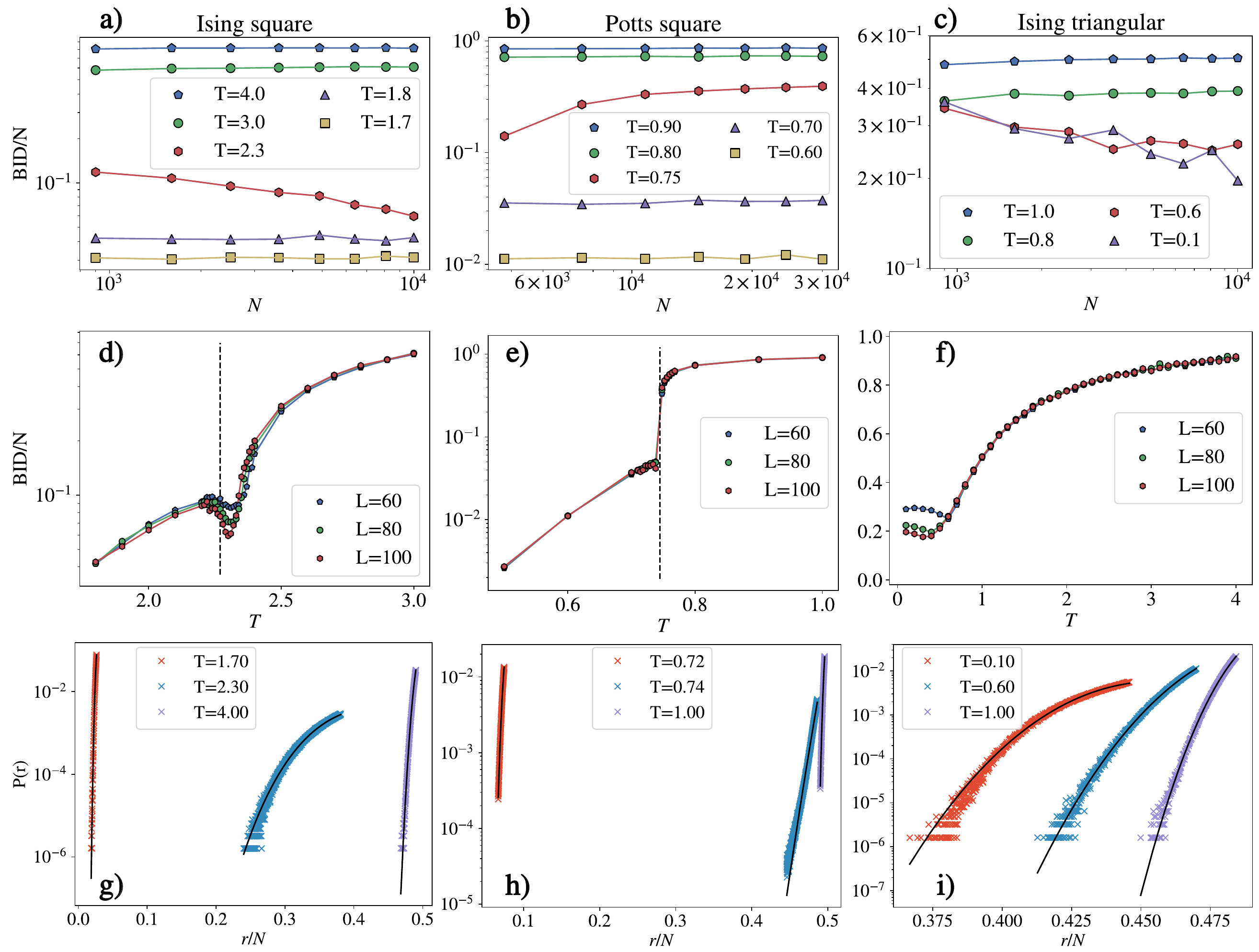}
\caption{
\textbf{The Binary Intrinsic Dimension (BID) of spin systems as models of interacting bit arrays.}
Upper row: BID per spin as a function of the total number of spins, $N$, for several temperatures $T$. 
Central row: Thermal dependence of the BID normalized per spin. $L$ stands for the lattice width (or height).
Bottom row: Model validation at $L=100$ and different temperatures.
In colored crosses, the empirical probability of Hamming distances $P_{emp}(r)$. In solid black line, our model fit, Eq. \eqref{eq:p-of-d}.
(a,d,g): ferromagnetic Ising model on the square lattice, 
(b,e,h): the Potts model on the square lattice with $q=8$ states, 
(c,f,i): the antiferromagnetic Ising model on the triangular lattice. 
With vertical dashed lines we show the exact results in the thermodynamic limit for the critical temperature, $T_c = 2/\log(1+\sqrt{2})$ in panel d) and the transition temperature $T^*=1/\ln{(1+\sqrt{q})}\approx 0.745$ in panel e).
}
\label{fig:spins-Nscaling}
\end{figure*}

To benchmark our approach we use spin systems from statistical physics (summarized in Fig. \ref{fig:spins-Nscaling}). In those systems, the variables are spins which can take two values (up or down) and which, therefore, are binary variables.  By choosing an interaction law between the spins, one can generate sets of binary variables correlated in highly non-trivial manners. The spins (or bits) are generated by sampling through standard Monte Carlo techniques\cite{newman1999monte} the  Boltzmann distribution $P(\boldsymbol{\sigma}) \propto \exp{(-\beta E(\boldsymbol{\sigma})})$, being $E$ the energy function encoding the interactions between spins and $\beta = 1/T$ the inverse temperature. 

We first show that our algorithm is able to compute intrinsic dimensions that scale linearly with size when the input data consists of statistically independent subsystems. We generated 2500 bit streams by concatenating $B$ blocks of $N=10^4$ bits, where each block is generated by independently harvesting spins from a Boltzmann distribution at  $T=2$, with an energy function $E(\boldsymbol{\sigma})=\sum_{i=1}^{N=1}\sigma_i \sigma_{i+1}$ with periodic boundary conditions ($\sigma_{N+1} \equiv \sigma_{1}$).  
Table \ref{tab:size-scaling} shows that the BID per bit is approximately constant from $10^4$ bits to $10^6$ bits, consistent with the system being constructed by concatenating independent blocks.  
Remarkably, the approach allows estimating intrinsic dimensions of order $10^5$ using only $2500$ data points. This is empirical evidence that the estimator proposed in this work is not affected by the curse of dimensionality, and can be used in extremely high dimensional spaces. For a detailed analysis of the dependence on the number of samples see Supplementary Note 2.

\begin{table}[ht]
\centering
\begin{tabular}{|c|c|c|c|c|}
\hline
Number of blocks $(B)$& 1          & 10       & 100 \\
\hline
Number of bits $(L)$  & $10^4$     & $10^5$   & $10^6$ \\
\hline
BID                   &  8047     & 80497  & 804624    \\ 
\hline
BID per bit           &  0.8047    & 0.8050 &  0.8046 \\
\hline
log(KL divergence)         &  -9.2      & -8.2     &  -7.1  \\
\hline
\end{tabular}
\caption{\textbf{BID linear scaling with system size for i.i.d concatenated blocks of bits.}
}
\label{tab:size-scaling}
\end{table}

Next, we show that the BID is capable of capturing correlations between bits even when those span very large distances. We consider spins harvested from the Boltzmann distribution of the two-dimensional ferromagnetic Ising model on the square lattice (see Methods, section `Spin systems' for details). 
At high temperatures, spin-spin correlations are short-range, namely, they decay exponentially fast with the lattice distance.
The characteristic scale of these correlations, the correlation length $\xi(T)$, is much smaller than the system size $L=\sqrt{N}$, implying that the BID must be asymptotically proportional to the number of bits. This is observed in the curve corresponding to $T=4.0$ in Fig. \ref{fig:spins-Nscaling}a). The value of the BID per bit is close to 1, reflecting the presence of small correlations. The same behavior is observed at $T=3$, with a relative BID that is smaller, consistently with the fact that correlations become more important.
At low temperatures, the system is globally ordered (namely the majority of bits are either $0$ or $1$) but the fluctuations are also short-range correlated\cite{binney1992theory}. Thus the BID is again proportional to the number of bits, but its value is much lower, reflecting the presence of a dominant strong correlation between the variables "forcing" most of the bits to align.
Near the critical temperature ($T=2.3$) the relative BID does not reach a plateau, indicating that the correlation length is comparable with the size of the system. This is a well-known phenomenon observed in spin systems at criticality\cite{binney1992theory}, which is captured by the BID through the sub-linear scaling with the number of bits.
Fig. \ref{fig:spins-Nscaling}d) shows the BID as a function of temperature for the 2D ferromagnetic Ising model. Away from criticality, the BID per spin tends smoothly to zero or one as the temperature goes to zero or infinity, respectively. The critical temperature is easily recognized in the graph since, consistently with ref. \cite{2NN-PT-Ising}, we observe a local minimum of the BID at that temperature.

The second system we studied is a $q$-state Potts model, with $q=8$ (see Methods, section `Spin systems' for details). 
In order to calculate its BID we construct a binary system by writing the state (or color) index $q$ in a binary representation. 
The correlation length in this model has a maximum at the transition temperature but remains finite even in the thermodynamic limit\cite{Potts-correlation-length}. This behavior is correctly captured by the BID scaling being asymptotically linear for every temperature (Fig. \ref{fig:spins-Nscaling}b), even close to the transition temperature $T^*$. 
The BID shows an abrupt change around $T^*$, consistent with the underlying first-order thermal phase transition.

Finally, we considered the Ising model on the triangular lattice with antiferromagnetic first-neighbor interactions. 
Due to geometrical frustration, this model does not have a thermal phase transition but presents a smooth crossover towards a ground state that has an entropy proportional to the number of spins in the system\cite{Wannier-triang} and moreover presents a power-law decay of correlations\cite{blote1993antiferromagnetic}.
Fig. \ref{fig:spins-Nscaling}c) captures the long-range correlations of the ground-state through the sublinear scaling of the BID per bit, only present at small temperatures. 
Decreasing the temperature, the BID per spin tends smoothly to a non-zero value for all system sizes studied (Fig. \ref{fig:spins-Nscaling}f). 

The bottom panels in Fig. \ref{fig:spins-Nscaling} illustrate the viability of the model defined in eq. Eqs \ref{eq:p-of-d} and \ref{eq:d-of-r} to describe the observed histograms of the distances between bit streams in all the conditions considered in the tests: the black lines, corresponding to the model of Eq. \ref{eq:p-of-d}, fit pretty accurately the observations in all the conditions, even if at different temperatures the observations are concentrated at very different distances.


\begin{figure*}[ht!]
\centering
\includegraphics[width=.8\linewidth]{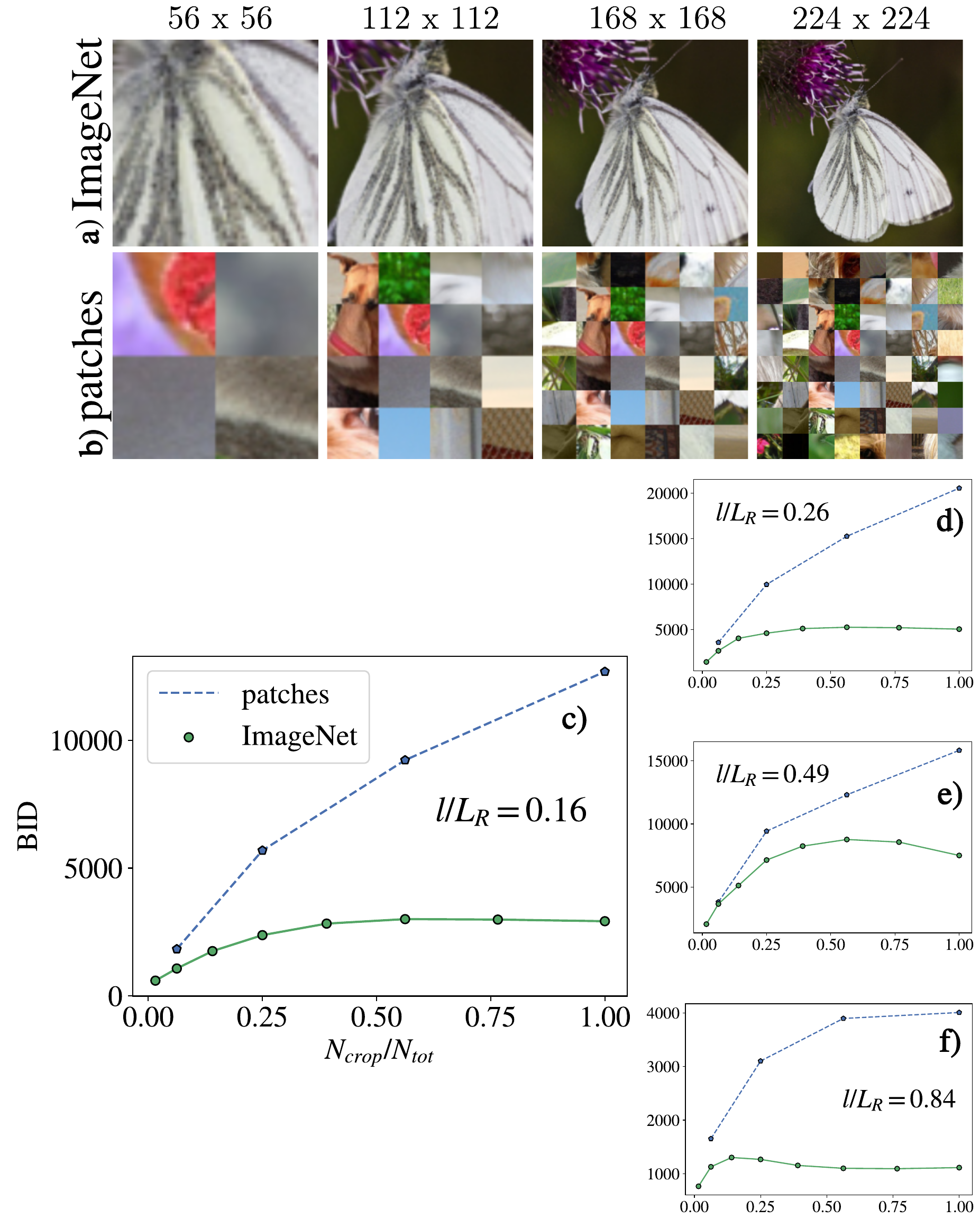}
\caption{\textbf{The Binary Intrinsic Dimension (BID) of image-crop representations in Resnet18.}
a) Example of crops for a sample of ImageNet. On top of each image, we report the size of the crop. All images are resized to $224 \times 224 $ resolution after cropping, thus the rightmost image is the full image. Note moreover that the crop is performed only along the spatial dimensions, keeping the three channels of the images.
b): a set of images constructed from random patches of size $28 \times 28 $ from ImageNet samples.
c), d), e), f) panels:  BID of image representations as a function of the number of pixels in the cropped image, $N_{crop}$, where we normalized the latter by dividing by the total number of pixels in the full image, $N_{tot}=224^2$. 
For each panel, $l$ is the layer index and $L_R$ is the total number of layers. 
For further details see Methods, section `The BID of image representations'. 
}
\label{fig:Resnet}
\end{figure*}

\begin{figure*}[ht!]
\centering
\includegraphics[width=.8 \linewidth]{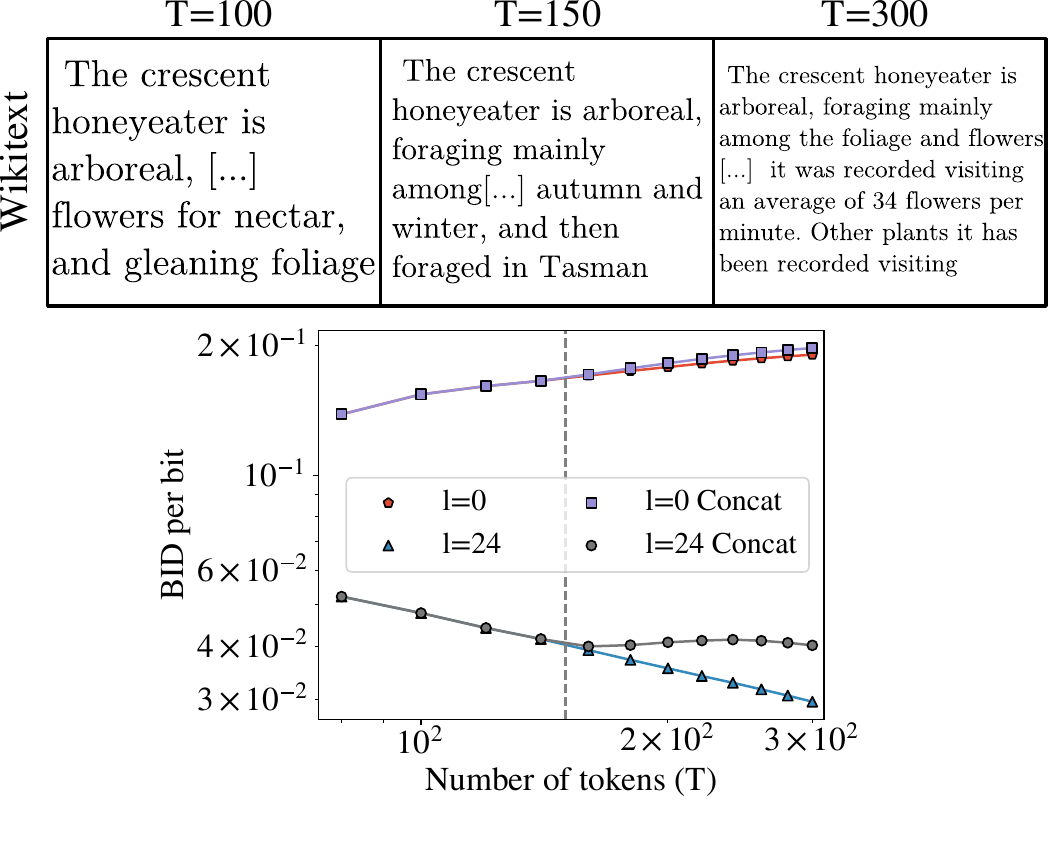}
\caption{\textbf{The Binary Intrinsic Dimension (BID) of text representations in large language models.} 
Upper boxes: an example of a Wikitext sentence truncated at different token lengths, $T$. 
Central panel: BID per bit calculated using all token representations in the sequence as a function of the number of tokens, for Pythia410m (see Methods, section `The BID of text representations' for details). 
$l=0$ stands for the embedding layer, $l=24$ corresponds to the last transformer layer before the head.
"Concat" stands for phrases constructed concatenating 150 tokens from two randomly selected data samples.
The dashed line corresponds to the semantic boundary between the two phrases.
}
\label{fig:text-fig}
\end{figure*}

These benchmark examples show that BID is able to characterize the collective behavior of a large number of bits.
Analogously, in computer vision, the collective behavior of a large number of pixels gives rise to patterns, which encode the meaning of an image. We computed the BID of internal representations of a convolutional image classifier, constructing binary data representations simply using the sign of the real-valued activations (see Methods, section `The BID of image representations' for details).
The BID behaves as a proxy of semantic content, as we show through the following size-scaling experiment.
Fig. \ref{fig:Resnet}a) shows an example of different crops for one element of ImageNet, where each crop is resized to the fixed full-image resolution.
Fig. \ref{fig:Resnet}c)-f) shows how the BID of the representations inside Resnet18 depends on the initial crop size, for different layers. 
For small crop sizes the BID increases monotonically, since on average the semantic content related to the classification task is being added.
For large enough crop sizes, the BID saturates, indicating that all the relevant content of the image was already present in previous sizes. 
Later layers saturate earlier since they carry information that is only relevant for the classification task and on average adding more context does not make significant changes.
Indeed, if we consider a sufficiently small crop of a generic image, it is not possible to assign a label to it, since it does not carry enough information to make a discrimination. Instead, semantic information emerges for big enough crop sizes but saturates when the crop is so large that the extra pixels do not bring new information. 
In Fig. \ref{fig:Resnet}c-f) we also show the same analysis performed on images constructed by an ensemble of random patches of the same data set. 
In this case, the semantic content is lost,  even if the local structure in the images is preserved. Consequently, the BID does not plateau, since large images correspond to a collection of independent crops, each bringing the addition of independent information. 

As a last example, we show that the BID can be used to detect correlations in the feature maps learned by large transformer-based architectures trained on a natural language processing (NLP) task.  
The ID has been used to analyze the properties of deep autoregressive transformers for protein sequence analysis and image analysis\cite{Valeriani2023} and more recently for language \cite{cheng2023bridging,cheng2024emergence}. 
However,  to make the ID estimation possible in Ref. \cite{Valeriani2023} the representations were averaged in the sequence dimension,  mixing information associated with the different tokens. In Refs.  \cite{cheng2023bridging,cheng2024emergence}, the analyses are performed only on the representation of the last token.
We here use our approach to estimate the binary intrinsic dimension of the representation of whole sentences, probing in this manner the collective properties of all the neurons of a layer. 
We consider paragraphs from Wikitext consisting of more than $300$ tokens, and we study how the BID scales with text length, binarizing activations through the application of the sign function.
The initial representation (layer 0, Fig. \ref{fig:text-fig}, left panel) has short-range correlations, signaled by the BID per bit slowly drifting up as a function of the number of tokens (Fig. \ref{fig:text-fig}). 
Token embeddings are correlated because they must be compliant with the grammar structure of natural language, in which words are often used in groups. For long sentences, the BID scales approximately linearly with the sentence length: grammar-related correlations become less relevant in parts of a sentence separated by hundreds of words. 
Instead, the BID per bit of the last-layer representation learned by the model displays a clear power-law decay as a function of the number of tokens, similar to the one observed in scale-free critical spin systems (see Fig. \ref{fig:spins-Nscaling}a).
Since the semantic content in text is a collective property of the whole sentence, the network to capture this property develops a long-range correlated representation: qualitatively the process of understanding the meaning of a long sentence can be described as the process of finding meaningful correlations in the sequence of tokens. 
To validate that we are indeed capturing the semantic correlations in text through the BID scaling we repeat the experiment by concatenating 150 tokens from two randomly chosen sentences from the same corpus. We observe that the power law decrease stops precisely at the boundary between phrases, and it is followed by an approximate plateau, consistent with the two halves of the paragraph containing independent information. 
In the last part of the plot, the  BID per bit starts decreasing again, as a result of the semantic correlations of the second half of the sentence. 
In Methods, section `The BID of text representations' we show that these results hold also for changing both corpus and network. 

These findings are aligned with pioneering studies that have suggested power-law correlations between characters in human text but were numerically limited due to the difficulty of computing mutual information in high dimensions\cite{li1989mutual,ebeling1994entropy}. 
Moreover, the critical behavior of activations complies with the more recent observation of neural scaling laws\cite{kaplan2020scalinglawsneurallanguage} where the loss function of autoregressive transformers decreases as a power-law of the number of parameters. 
Also consistent with our results, a recent work observed a critical phase transition in GPT-2, with the temperature parameter appears in the autoregressive sampling\cite{nakaishi2024critical}. To make computations feasible, the authors restricted the representation space to that of Part-Of-Speech (POS) tags only.

\section*{Comparison with other methods}
\label{sec:results-comparison}

\begin{figure}[H]
\centering
\includegraphics[width=\linewidth]{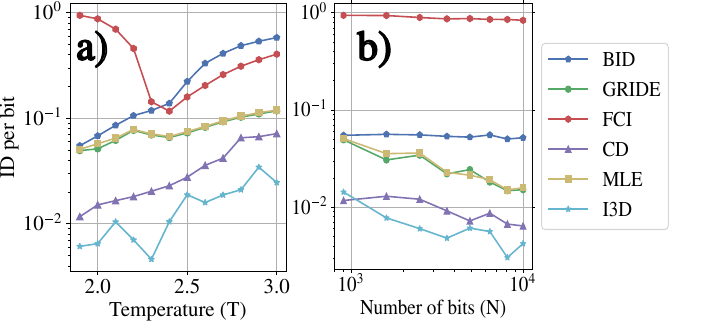}
\caption{\textbf{Comparison between Binary Intrinsic Dimension (BID) and other real-space estimators.}
Panel a): Thermal dependence of the intrinsic dimension for a small $30 \times 30$ ferromagnetic Ising model.
Panel b) scaling of the intrinsic dimension normalized per bit at temperature $T=1.9$. 
Number of samples: 5000.
GRIDE stands for the Generalized Ratios Intrinsic Dimension Estimator\cite{denti2022generalized}.
FCI stands for the Full Correlation Integral estimator\cite{erba2019intrinsic}.
CD stands for Correlation Integral\cite{grassberger1983measuring}.
MLE stands for the Maximum Likelihood Estimator of Ref. \cite{levina2004maximum}.
I3D stands for the Intrinsic Dimension Estimator for Discrete Datasets of Ref. \cite{I3D}.
}
\label{fig:comparisons-Ising2D}
\end{figure}

We compared our results with those obtained with other intrinsic dimension estimators, in particular: (i)  the Correlation Dimension\cite{grassberger1983measuring} (CD), perhaps the simplest and most widely know approach for estimating the ID; (ii) the MLE estimator  in Ref. \cite{levina2004maximum} and GRIDE\cite{denti2022generalized},  which estimate the ID by maximizing the likelihood of the empirical probability of the ratio of the distances of the nearest neighbors. (iii) I3D\cite{I3D}, which is designed to estimate the intrinsic dimension in data spaces in which the features take discrete values, and (iv) FCI\cite{erba2019intrinsic}, 
an ID estimator that was shown to infer the correct number of dimensions for up to 30 non-interacting spins and also for toy distributions in dimensions of the order of hundreds.

First, for a small 30x30 ferromagnetic Ising model we studied the thermal dependence of the ID (Fig. \ref{fig:comparisons-Ising2D}-a)).
At low temperature the system is in the ferromagnetic phase, which implies that a large fraction of the spins points in the same direction.
We observe for all estimators except FCI that in these conditions the estimated ID is small and of roughly comparable magnitude. The estimates are consistent because in this special regime the local estimators (MLE, GRIDE, I3D and CD) are not severely affected by the dimensionality.
In the same regime, the  FCI estimator predicts a qualitatively different behavior, with the ID being very high and a decreasing function of T. This is at odds with physical intuition.  Indeed, in a Ising model at low temperature only few defects are present, and  the state of the system can be fully described by providing only the position of those defects. Therefore, the ID should smoothly tend to zero for $T\rightarrow 0$, similarly to the entropy.
We notice that in the ferromagnetic phase most of the spins have the same value. In these conditions the centering and normalization procedure at the basis of FCI places the data on a hypersphere, but the distribution of the data on it is highly non-uniform. The qualitatively different results obtained by FCI might then be induced by the violation of one of its assumptions, that  the data are  isotropic when observed from the geometric center of the data set.
In the paramagnetic phase, the BID estimate is instead comparable to the FCI estimate. Indeed, in this phase the spins are not polarized, namely the magnetization defined as $M=\sum_{i=1}^N \sigma_i$ is null. In these conditions, after centering and normalization, the data will be approximately isotropic, making the estimate by FCI more reliable. 
In contrast, all the other local estimators subestimate the value of the ID because they need an exponential  number of samples in order to converge.

Fig. \ref{fig:comparisons-Ising2D}-b) shows the size scaling of the different algorithms at fixed low temperature, where we know the system is extensive, and thus the ID must scale linearly with the number of bits, $N$.
For small system sizes we observe that the local estimators (MLE, GRIDE, I3D and CD) give results comparable with the BID, but their estimate of the ID does not scale linearly with $N$, at odds with physical evidence. This is the behavior which prompted us to develop a new estimator, specifically designed for bits or spins. The BID scales linearly with the number of bits, and so does FCI. Nonetheless for FCI the slope of the line is one order of magnitude bigger and, as discussed above, the corresponding estimation is also at odds with physical intuition.

We finally performed ID estimates with GRIDE and FCI on the images and text databases.
For our experiment on images, we observe that both GRIDE and FCI behave qualitatively similarly to the BID, namely the plateau observed for real data and its absence for images constructed grouping random patches are reproduced (see Supplementary Note 3.1). 
However, we remark that the ID estimation performed with the GRIDE is  more than one order of magnitude smaller than the estimations done by BID, whereas the estimations of FCI are systematically below that of BID, by roughly a factor of 2.
Additionally, in Supplementary Note 3.3  we study the BID of simple images directly in pixel space, for MNIST and CIFAR10. There we show that the BID of MNIST is fully compatible with other real-space estimators, and we discuss the challenges of finding correct binarizations for RGB images.
For the text experiment, we show in Supplementary Note 3.2 that GRIDE fails to display a plateau at the semantic boundary since, as we showed in Fig. \ref{fig:comparisons-Ising2D}, it doesn't scale correctly with system size. 
Instead, the FCI estimator produces qualitatively comparable results to ours, reproducing the linear scaling for the ID at the layer zero and detecting the semantic boundary in the case of concatenated paragraphs.
Nonetheless, we observed significant differences in the size scaling exponents of the BID normalized per bit at the last layer of the transformer, namely the slope $\eta$ of the lines in the log-log plot of $BID(T)/N \propto T^{-\eta}$. Whereas for the BID we get an exponent of $ \eta_{BID} = 0.42$, for FCI and GRIDE we get much higher values, $\eta_{FCI} = 0.89$, and $\eta_{GRIDE}=0.70$.
The study of the relationship between the BID and the scaling exponents in systems with power-law correlations, and in particular in critical systems, together with the quantitative discrepancies observed, are interesting topics which we are currently investigating.

Wrapping up, if the ID is small, our estimator is qualitatively consistent with all the other estimators we considered, except FCI if  the bits have a preferential value (like in a ferromagnetic phase). 
When the ID is large, our estimator provides values closer to the FCI estimator, while the other estimators provide lower values. 
Only our estimator and the FCI estimator provide ID estimates which scale linearly with system size, although in Fig. \ref{fig:comparisons-Ising2D} we observed that the FCI values of the ID per spin are almost one order of magnitude larger in the ordered phase than in the disordered phase, a result that is not plausible.

The significant quantitative differences with  respect to MLE, CD, I3D, GRIDE and FCI observed in some regimes can be understood by considering that the estimate of the ID depends on the assumption on the geometrical structure of the manifold containing the data. For binary data this manifold is a N-dimensional hypercube, and the data can only occupy its vertices. The I3D estimator assumes that the manifold is an ID-dimensional cubic lattice.  The resulting estimator correctly captures the dimension of points with discrete-valued features when the intrinsic dimension is moderate. However, the difference in the geometric structure of the manifold (a cube of size one for BID, a lattice of infinite size for I3D) determines qualitative differences when the ID is large. The CD, MLE, GRIDE and FCI are all implicitly or explicitly based on the assumption that the features are real numbers. In FCI it is also explicitly assumed that the data are isotropic, namely that that are uniformly present in all the directions with respect to the geometric center of the dataset. These assumptions do not hold, even approximately, in binary data, bringing to systematic errors.


\section*{Conclusions}
We introduced an intrinsic dimension estimator specifically designed for binary data. To the best of our knowledge, the only other estimator for such variables was recently introduced in the context of formal concept analysis \cite{hanika2024textitintrinsicdimensionbinarydata}. However, it has only been benchmarked on simple binary data tables, whereas our approach is tailored for large bit streams. We tested examples with up to $10^6$ bits without observing systematic errors in the scaling of the estimation with system size.

The spin system benchmarks demonstrate that our estimator can characterize the global correlation structure of different phases of matter and identify the corresponding phase transitions. 
When applied to data representations from deep neural networks, the scaling of BID allows us to infer correlations for both image recognition and language modeling tasks. 
In the case of image classification, the BID behaves as a proxy of semantic content, since it initially increases with image-crop size after which it plateaus, indicating no additional information is gained from extra pixels. Instead, for images constructed out of independent random patches this plateau does not take place. 
For sentences, the correlations extend throughout the text and the BID of the last transformer layer follows a power law when computed using representations from a real corpus. In contrast, in an artificial corpus generated by concatenating real sentences, these correlations break, showing that the power law is inherent to the data and not an artifact of our estimator.

It is noteworthy that the last two datasets consist of real-valued features, while the BID estimator is designed for binary data. Each feature was converted into a binary variable by taking its sign. In Methods, section `The BID of text representations' we show that the trends observed in Fig. \ref{fig:text-fig} hold when using two bits per feature instead of one. Specifically, the relative BID estimated for this higher-precision representation scales with the same power law. This suggests that when the intrinsic dimension becomes extremely large (around 1000 or more), its trends are insensitive to the precision used to represent individual features, and binning retains the essential information of the representation, as suggested in Refs. \cite{hubara2018quantized,courbariaux2016binarized}.


The BID of data representations is clearly not restricted to either images or text, which are taken here as use cases where semantic correlations are easily interpretable. 
We expect our algorithm to be useful  in the analysis of generic high dimensional spaces where global correlations are unknown, for example data coming from stock markets or brain activity. 
There are a number of interesting perspectives, among which we can mention understanding the influence of different binarization protocols of generic continuos data on the resulting BID; the use of analytical solutions in simple models for the distribution of Hamming distances, $P(r)$, to gain additional physical insights on the respective BID; and exploring the relationship between the BID and the Shannon entropy estimated by compression algorithms\cite{avinery2019universal}.

\section*{Code availability}
The code used to obtain the results of this work can be found in \href{https://github.com/acevedo-s/BID}{https://github.com/acevedo-s/BID}. Routines to compute the BID are implemented in the open-source package DADApy\cite{GLIELMO2022100589}. A tutorial notebook can be found at 
\href{https://github.com/sissa-data-science/DADApy/blob/main/examples/notebook_hamming.ipynb}{https://github.com/sissa-data-science/DADApy/blob/main/examples/notebook\_hamming.ipynb}

\section*{Data availability}
The datasets generated during and/or analysed during the current study are available from the corresponding author on reasonable request.

\section*{Acknowledgements}
The authors acknowledge Marco Baroni, Antonello Scardicchio and Marcello Dalmonte for useful discussions.
AL and SA acknowledge foundings from  NextGenerationEU through the Italian National Centre for HPC, Big Data, and Quantum Computing (Grant No. CN00000013).
AL and SA also acknowledge financial support by the region Friuli Venezia Giulia (project   F53C22001770002)

\section*{Methods}

\subsection*{Bayesian inference of the BID by Markov Chain Monte Carlo}
\label{sec:Bayes}

\begin{figure}[H]
\centering
\includegraphics[width= \linewidth]{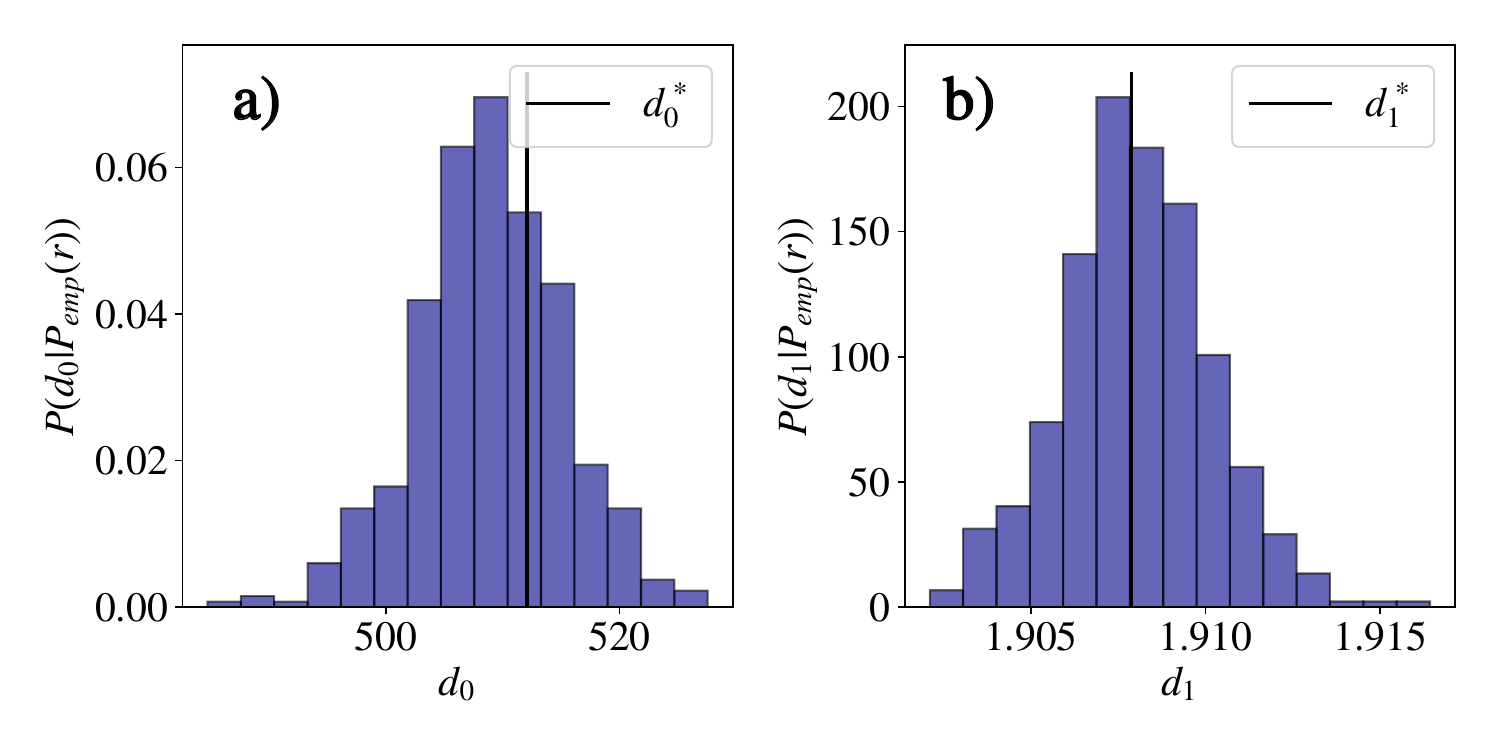}
\caption{ \textbf{BID Bayesian inference.}
Panels a) and b) show the empirical posterior densities for the two parameters of our model, $d_0$ and $d_1$, respectively.
$d_0^*/N=0.0512$ and $d_1^*=1.9079$ correspond to the solution found optimizing Eq. \eqref{eq:p-of-d}. The posterior means and standard deviations are $\mu_{d_0}/N = 0.0510$, $\sigma_{d_0}/N=0.0006$, $\mu_{d_1}=1.908$, $\sigma_{d_1}=0.002$. Number of samples: $N_s = 500$, number of spins $N=10^4$, temperature $T=2.3.$}
\label{fig:Bayes}
\end{figure}
The parameters of model \eqref{eq:p-of-d} can be found minimizing Eq. \eqref{eq:KLdiv}, as we did in our work, or alternative using the Metropolis Hastings algorithm on model \eqref{eq:p-of-d}, with a flat prior. Here we sample the probability distribution
The parameters of model \eqref{eq:p-of-d} can be found minimizing Eq. \eqref{eq:KLdiv}, as we did in our work, or alternative using the Metropolis Hastings algorithm on model \eqref{eq:p-of-d}, with a flat prior. Here we sample the probability distribution
\begin{equation}
    P(d_0,d_1|P_{emp}(r)) = \prod_{r \in \text{data}} \bigg ( \frac{\binom{d(r)}{r}}{2^{d(r)}} \bigg)^{N_r},
\end{equation}
where $N_r$ is the number of times the distances $r$ was sampled in the dataset.
Fig. \ref{fig:Bayes} shows the equivalence of the two methods on the 2D ferromagnetic Ising model at $T=2.3$. 
The MCMC simulation was performed starting at $(d_0,d_1) = (\langle r \rangle_{P_{emp}},1)$, making random moves with a maximum amplitude given by 1 percent of the current value of $(d_0,d_1)$ at each time step $t$. This approach allows estimating the mean, variance and covariance of the parameters of the model.

\subsection*{Spin systems}
For every two dimensional spin model we sampled $5000$ samples at each temperature with singly spin flip dynamics. We simulated each of the $5000$ Markov chains independently in a different core with a different seed, so they are by construction independent from each other. 
For Ising models the algorithm used is Metropolis-Hastings, whereas for the Potts model we used the Heatbath algorithm\cite{newman1999monte}. 
For Table \ref{tab:size-scaling} we used 2500 samples. 
\subsubsection*{Ising models}
For all the Ising models simulated in this work, the energy of a given configuration  of $N$ spins
$\boldsymbol{\sigma}=(\sigma_1,...,\sigma_N) \in \{-1,1\}^N$ is
\begin{equation}
E(\boldsymbol{\sigma}) = -J\sum_{\langle i,j\rangle} \sigma_i \sigma_j,
\label{eq:Ising-hamiltonian}
\end{equation}
with periodic boundary conditions, where $\langle i,j\rangle$ stands for first-neighbor interactions between spins $i$ and $j$ in the corresponding lattice. $J$ Is the coupling constant between spins, fixed to $+1$ in the ferromagnetic models and $-1$ in the antiferromagnetic case. 

\subsubsection*{Potts model}

The energy of a given configuration $\boldsymbol{\sigma}=(\sigma_1,...,\sigma_N) \in \{0,1,...,q-1\}^N$ is
\begin{equation}
    E(\boldsymbol{\sigma}) = -\sum_{\langle i,j\rangle} \delta_{\sigma_i,\sigma_j},
    \label{eq:potts-hamiltonian}
\end{equation}
with periodic boundary conditions, where $q \in \mathds{N}$ and $\delta$ stands for Kronecker delta. For $q>4$ the system presents a first order phase transition. 
Since in this model all colors are equivalent, to improve the statistics we made a symmetry transformation making all simulations converge to the $0-$ground state. 

\subsection*{The BID of image representations}
\label{sec:A_Resnet}

We followed the experimental setup of \cite{ansuini2019intrinsic} by computing separately the BID of Resnet18 representations from 8 highly populated classes of ImageNet ('vizsla', 'koala', 'Shih-Tzu', 'Rhodesian\_ridgeback', 'English\_setter', 'cabbage\_butterfly', 'Yorkshire\_terrier'), where for each class we have roughly 1300 images. We measure the BID after each Resnet block, i.e., after each skip connection. Resnets have ReLU activation functions, so only two possible outputs of each unit are possible: zero or greater than zero. Taking the sign of such an output automatically constructs the binary system under the convention $sign(0)=0$.
The preprocessing of images was done following PyTorch official documentation.
The layer indexation was taken following the list of "graph\_node\_names", from the official PyTorch documentation.

\subsection*{The BID of text representations}

\begin{figure}[H]
\centering
\includegraphics[width=.75\linewidth]{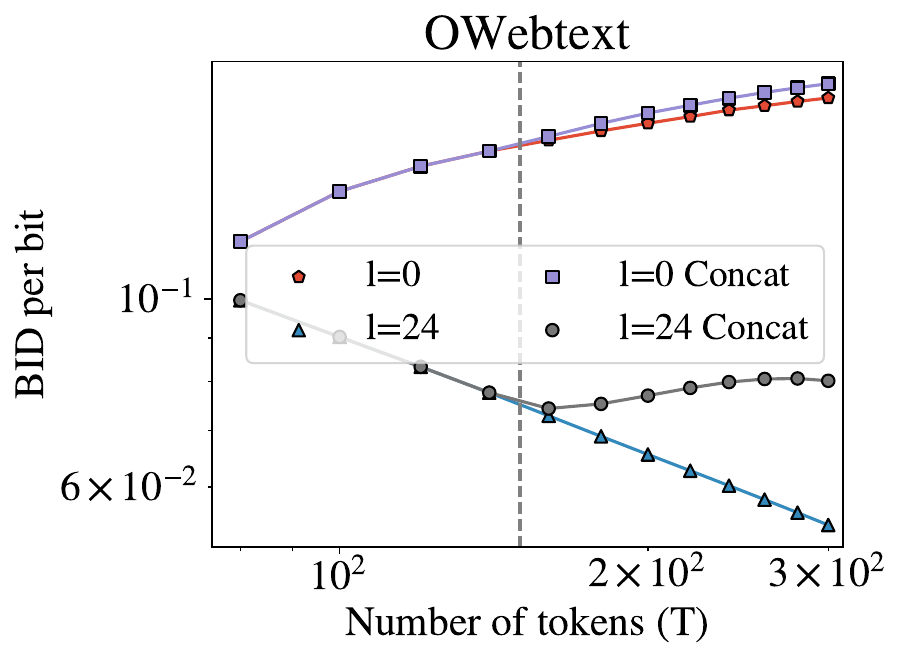}
\caption{\textbf{Stability of results against changes in both data and model.}
BID per bit calculated using all token representations in the sequence as a function of the number of tokens, for OPT350m on OpenWebtext. 
l = 0 stands for the embedding layer, l = 24 corresponds to the last transformer layer before the head. "Concat" stands for phrases constructed concatenating 150 tokens from two randomly selected data samples.
The dashed line corresponds to the semantic boundary between the two phrases.}
\label{fig:OPT-OpenWebtext}
\end{figure}

Pretrained LLMs OPT350m and Pythia450m and associated tokenizers where downloaded from Hugginface\cite{wolf2019huggingface}, together with Wikitext and OpenWebtext corpora. Model architectures are available online. 
The activation function in the last fully connected layer of the transformer block in Pythia is GeLU, whereas in OPT the corresponding layer is linear (i.e., no activation function).
Pythia450m has layer\_norm before and after the self attention layer, whereas OPT350m has it between the attention block and the fully connected layers. 
Regardless of these details the binarization corresponds to take the sign of the activations (which in every layer have mean zero), and we see no qualitative difference when comparing results from Pythia on Wikitext to OPT on OpenWebtext (Fig. \ref{fig:OPT-OpenWebtext}). 
Discarding sentences from Wikitext with less than 400 Tokens we kept around 7500 samples. 

To represent each activation of Pythia410m on Wikitext with 2 bits we proceed as follows. For each layer $l$ we computed the mean $\mu^l$, which is close to zero, and the standard deviation $\sigma^l$ of the activations $a_i^l$, with $i=1,...,N_a^l$ being $N_a^l$ the number of activations in layer $l$. Then, we quantize the activations generating 2-bit variables $\sigma_i^l$ by 

\begin{equation}
    \sigma_i^l = \begin{cases}
        00 \quad \text{if} \quad  a_i^l < - \sigma^l \\
         01 \quad \text{if} \quad 
        - \sigma^l < a_i^l < 0 \\
        10 \quad \text{if} \quad 
        0 < a_i^l <  \sigma^l\\
        11 \quad \text{if} \quad 
           \sigma^l < a_i^l\\
    \end{cases}
    \label{eq:2-bit-representation}
\end{equation}

\begin{figure}[H]
\centering
\includegraphics[width=.75\linewidth]{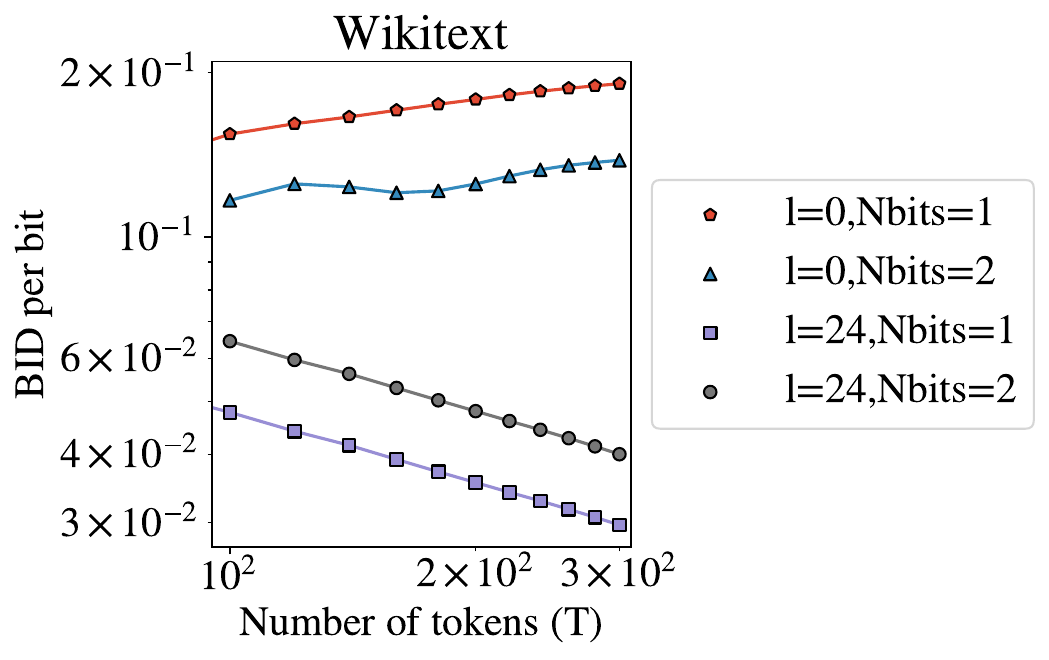}
\caption{\textbf{Stability with respect to binarization protocol}.
Comparison of BID size scalings using  direct sign binarization (Nbits=1) or the binarization given by \eqref{eq:2-bit-representation}.}
\label{fig:bits-comparison}
\end{figure}

Fig. \ref{fig:bits-comparison} shows the comparison between sign binarization (Nbits=1) and the quantization \eqref{eq:2-bit-representation} (Nbits=2).

\subsection*{Maximum Likelihood Estimation of a scale-dependent BID}

In this section we present a formal approach to define and estimate a scale-dependent BID, $d(r)$, that supports the ansatz \eqref{eq:p-of-d} with the functional form \eqref{eq:d-of-r}. 
Following \cite{I3D}, we assume to have a Poisson process\cite{MOLTCHANOV2012Poisson} in a generic domain in which we can define a set $A$ and assign to it a volume, $V_A$. Assuming also to have a constant density of points $\rho$ in $A$, the probability of observing $n_A$ points falling inside $A$ is 
\begin{equation}
 P(n_A,A) = \frac{(\rho V_{A})^{n{_A}}}{n_A!} e^{-\rho V_{A}},
\end{equation}
such that $\langle n_A \rangle = \rho V_{A}$.
We consider a data point $i \in A$, and another set $B$ that contains $A$. Then, if the density is constant in set $B$ ($\rho_A = \rho_B$ = $\rho$) the conditional probability of having $n_A$ points in $A$ given that there are $n_B$ points in $B$ is

\begin{equation}
P(n_A | n_B) = \binom{n_B}{n_A}p^{n_A} (1-p)^{n_B-n_A},
\end{equation}
where $p \equiv V_{A}/V_{B}$. Note that this conditional probability does not depend on the density of points $\rho$. Then, the likelihood function is 

\begin{equation}
 \mathcal{L}(n_A^i|n_B^i,p) = \prod_{i=1}^{N_s} Binomial(n_A^i|n_B^i,p),
\end{equation}
where $N_s$ is the number of samples in the dataset, and identical volumes $V_A$ and $V_B$ where taken around every data point $i$. This equation is maximized to obtain the Maximum Likelihood Estimator $\hat{d}$. Doing this, in the limit of large number of samples, the equation that defines $\hat{d}$ takes the form

\begin{equation}
 \frac{V_A(\hat{d})}{V_B(\hat{d})} = 
 \frac{\langle n_A \rangle}{\langle n_B \rangle},
 \label{eq:d-MLE}
\end{equation}
where $\langle n_A \rangle = \frac{1}{N_s} \sum_{i=1}^{N_s} n_A^i$, and the same for $\langle n_B \rangle$.
To work specifically with Ising spin configurations $\boldsymbol{\sigma} \in \{ -1,1\}^N$ we depart from \cite{I3D} using Hamming distances, which can be defined as 

\begin{equation}
D_H(\boldsymbol{\sigma},
\boldsymbol{\sigma}') = 
\frac{N - \boldsymbol{\sigma}\cdot \boldsymbol{\sigma}'}{2}
\end{equation}

The number of points at Hamming distance $r$ from any Ising spin configuration in $d$ dimensions is 

\begin{equation}
\begin{split}
 V_H(r,d) &= \sum_{r'=0}^r\binom{d}{r'} 
\end{split}
\end{equation}
where $r \in [0,d]$. At this point we have to choose specific sets $A$ and $B$ to estimate $d$. To favor the constant density $\rho$ and constant dimension $d$ on both sets, one can fix $A$ to be the set of points at exactly distance $r_A$ from the data point $i$, and $B$ the set of points at distance $r_A$ or $ r_B = r_A+1$. We denote this two sets as "thin shells". Given this choice, \eqref{eq:d-MLE} has the close  form solution

\begin{equation}
\frac{V_A}{V_B} = \frac{r_B}{d+1},
\label{eq:p_thin_shells}
\end{equation}
for which the MLE for the ID is 

\begin{equation}
\hat{d}(r_A) = 
 \bigg ( 
 (r_A+1) \frac{\langle n_B \rangle}{\langle n_A \rangle}
 \bigg ) -1.
\end{equation}

In order to interpret better the last equation, we observe that $\langle n_A \rangle \propto P(r=r_A)$, where $P(r)$ is the probability of observing hamming distance $r$ between any two configurations in the dataset. Then, 

\begin{equation}
\begin{split}
\hat{d}(r_A)&\approx(r_A+1)\frac{P(r=r_A)+P(r=r_A+1)}{P(r=r_A)}-1 = \\
& \approx
(r_A+1)\bigg ( 
2+\partial_r \log{(P(r))} \biggr\rvert_{r=r_A}\bigg)-1.
\end{split}
\label{eq:BID-as-a-slope}
\end{equation}
where in the second step we did a first order Taylor expansion. Thus, given the empirical $P(r)$ this setup allows to estimate a scale-dependent  intrinsic dimension from the slope of the log-probabilities at a given scale $r_A$. 


\begin{figure}[t]
    \centering
    \includegraphics[width=\linewidth]{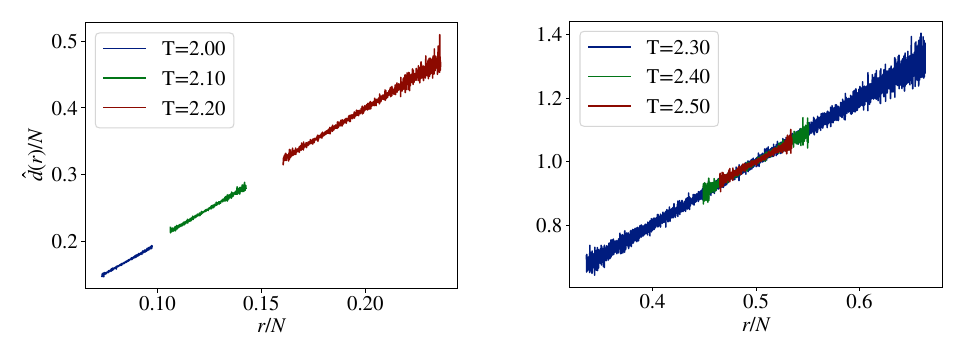}
    \caption{ \textbf{Maximum likelihood estimation of local BID.}
Scale-dependent BID estimated by MLE \eqref{eq:d-MLE} per bit, $\hat{d}(r)/N,$ as a function of the observational scale $r$, on the two-dimensional ferromagnetic Ising model on the square lattice, for several temperatures $T$. 
    Both axis are normalized by the total number of spins $N$ in the system. The two panels are separated temperatures above (right) and below (left) the critical temperature of the system for visual clarity. 
    }
    \label{fig:MLE-BID}
\end{figure}



Fig. \ref{fig:MLE-BID} shows the scale dependence of the MLE \eqref{eq:BID-as-a-slope} for the 2D Ising model where we observe a linear scale dependence in the whole range of probed scales, for every temperature. This linear dependence is consistent with  the success of our ansatz \eqref{eq:p-of-d} with a scale dependent $d(r)$ truncated to linear order (Eq. \eqref{eq:d-of-r}).
The scale-dependent BID \eqref{eq:d-MLE} presents  some drawbacks which are solved by the ansatz \eqref{eq:p-of-d}. First of all, there is no clear scale to select in either panel of Fig. \ref{fig:MLE-BID}, whereas by construction the coefficients of expansion \eqref{eq:d-of-r} are scale independent. 
Furthermore, at high temperature (right panel of Fig. \ref{fig:MLE-BID}), for scales greater than $N/2$ the BID per bit becomes larger than 1, which is a paradox. 
We also observed this linear dependence of the BID as a function of the observational scale $r$ studying ImageNet representations inside a Resnet. 
The results were noisy, since the MLE estimator uses the slope of the empirical log probabilities at a given value of $r$. 
This led us to define our ansatz in the form of Eqs. \eqref{eq:p-of-d} and \eqref{eq:d-of-r} which is not noisy, since it fits the empirical histogram in a large range of scales. 

\section*{References}
\bibliography{refs}


%
\section*{Author contributions}
SA performed the numerical experiments. SA, AR, and AL designed the experiments, analyzed the data, and wrote the manuscript.
\section*{Competing interests}
The authors declare no competing interests.

\section*{Supplementary Material}
Supplementary Material is available in the \href{https://www.nature.com/articles/s42005-025-02115-z}{published open access version}, Ref.~\cite{acevedo2025unsupervised}.

\end{document}